\documentclass[sigconf]{acmart}

\copyrightyear{2025}
\acmYear{2025}
\setcopyright{cc}
\setcctype{by}
\acmConference[WWW Companion '25]{Companion Proceedings of the ACM Web Conference 2025}{April 28-May 2, 2025}{Sydney, NSW, Australia}
\acmBooktitle{Companion Proceedings of the ACM Web Conference 2025 (WWW Companion '25), April 28-May 2, 2025, Sydney, NSW, Australia}
\acmDOI{10.1145/3701716.3717568}
\acmISBN{979-8-4007-1331-6/2025/04}

\usepackage{graphicx} 
\usepackage{booktabs} 
\usepackage{algorithm}
\usepackage{algorithmic} 
\usepackage{svg} 
\usepackage{listings} 

\usepackage{amssymb}
\usepackage{balance}

\title{Are Large Language Models Good Data Preprocessors?}
\author{Elyas Meguellati}
\affiliation{%
  \institution{The University of Queensland}
  \city{Brisbane}
  \state{Queensland}
  \country{Australia}
}
\email{m.meguellati@uq.edu.au}

\author{Nardiena Pratama}
\affiliation{%
  \institution{The University of Queensland}
  \city{Brisbane}
  \state{Queensland}
  \country{Australia}
}
\email{uqnp@uq.edu.au}

\author{Shazia Sadiq}
\affiliation{%
  \institution{The University of Queensland}
  \city{Brisbane}
  \state{Queensland}
  \country{Australia}
}
\email{s.sadiq@uq.edu.au}

\author{Gianluca Demartini}
\affiliation{%
  \institution{The University of Queensland}
  \city{Brisbane}
  \state{Queensland}
  \country{Australia}
}

\let\balance\relax
\begin{document}

\begin{abstract}
High-quality textual training data is essential for the success of multimodal data processing tasks, yet outputs from image captioning models like BLIP and GIT often contain errors and anomalies that are difficult to rectify using rule-based methods. While recent work addressing this issue has predominantly focused on using GPT models for data preprocessing on relatively simple public datasets, there is a need to explore a broader range of Large Language Models (LLMs) and tackle more challenging and diverse datasets.
In this study, we investigate the use of multiple LLMs, including LLaMA 3.1 70B, GPT-4 Turbo, and Sonnet 3.5 v2 to refine and clean the textual outputs of BLIP and GIT. We assess the impact of LLM-assisted data cleaning by comparing downstream-task (SemEval 2024 Subtask `Multilabel Persuasion Detection in Memes'') models trained on cleaned versus non-cleaned data. While our experimental results show improvements when using LLM-cleaned captions, statistical tests reveal that most of these improvements are not significant. This suggests that while LLMs have the potential to enhance data cleaning and repairing, their effectiveness may be limited depending on the context they are applied to and the complexity of the task and the level of noise in the text. Our findings highlight the need for further research into the capabilities and limitations of LLMs in data preprocessing pipelines, especially when dealing with challenging datasets, contributing empirical evidence to the ongoing discussion about integrating LLMs into data preprocessing pipelines.
\end{abstract}



\maketitle

\section{Introduction}

\begin{figure*}[t]
    \centering
    \includegraphics[width=\linewidth]{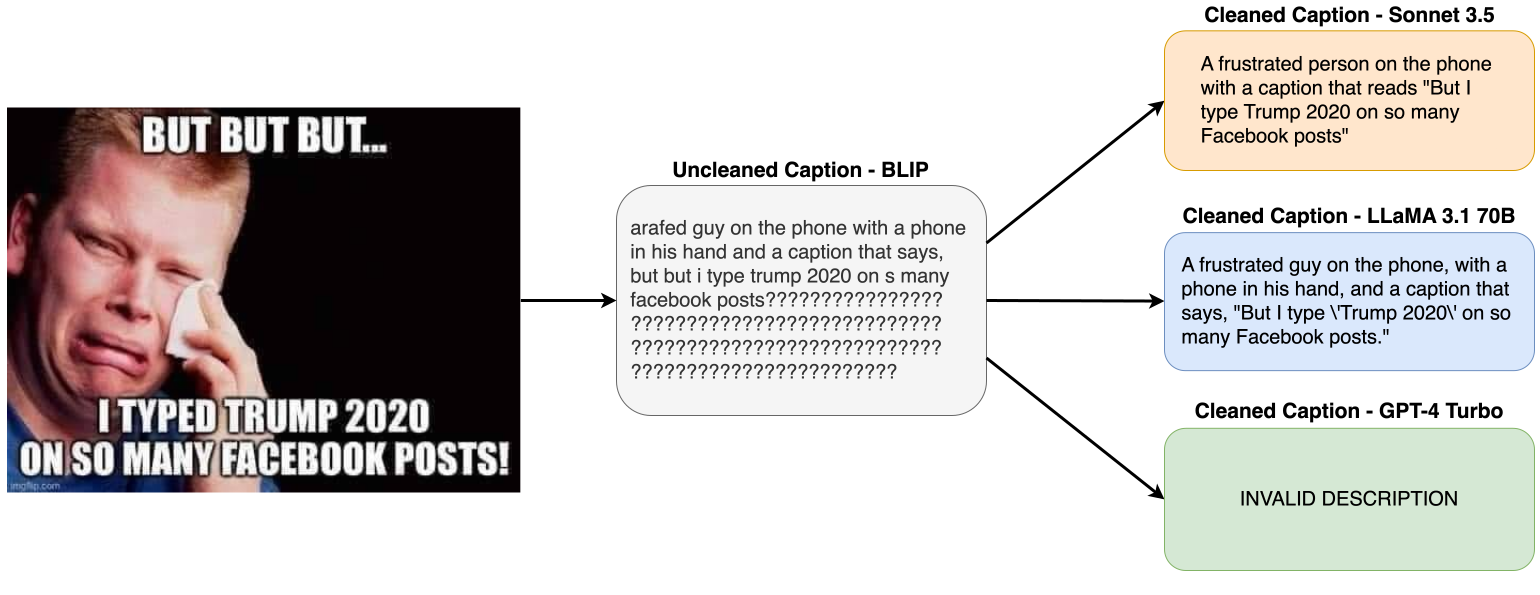}
    \caption{Caption Cleaning Process - Before and After. The original caption is cleaned using three different Large Language Models: Sonnet 3.5, LLaMA 3.1 70B, and GPT-4 Turbo.}
    \Description{This figure illustrates the caption cleaning process using three different Large Language Models. It shows a comparison of the original caption with the cleaned versions.}
    \label{fig:caption_cleaning_process}
\end{figure*}

Large Language Models (LLMs) have demonstrated remarkable versatility across diverse domains, from code generation and creative writing to complex reasoning tasks and data analysis. Their ability to understand context, generate human-like text, and adapt to various tasks through prompt engineering has led to their integration into numerous applications. Recent research has shown promising results in using LLMs for specialized tasks such as SQL query generation \citep{zhong2023sql}, data wrangling \citep{rahman2023data}, and automated data analysis \citep{liang2023llmdata}. The applications of LLMs have expanded into scientific domains, with successful implementations ranging from fine-tuned GPT-3 models answering complex questions in chemistry and materials science \citep{jablonka_leveraging_2024} to multimodal LLMs learning from unlabelled medical images and text \citep{liu_medical_2023}. However, one crucial area that remains relatively unexplored is their potential role in improving the quality of outputs from other AI systems.

The rapid advancement of multimodal AI systems has driven the widespread adoption of image captioning models such as BLIP and GIT \citep{li2022blip, wang2022git}. Despite their sophistication, these models often produce outputs that exhibit inconsistencies, errors, and contextual misalignments, which pose significant challenges for downstream NLP tasks. Traditional methods for cleaning such outputs have predominantly relied on rule-based systems. While these methods can address specific error patterns, they frequently fall short in capturing the subtle and diverse inaccuracies characteristic of AI-generated text. Recent studies have begun to explore the potential of LLMs in data preprocessing tasks \citep{zhang2024largelanguagemodelsdata, zhang2024jellyfishlargelanguagemodel}; however, their capability to effectively clean and refine outputs from other AI systems remains an underexplored area.

This study investigates whether LLMs can effectively preprocess noisy textual data, specifically focusing on captions generated by automated image captioning models. We explore this question through the lens of a particularly challenging downstream task: multilabel persuasion detection in memes. The motivation behind leveraging automated captioning for persuasive meme classification lies in the inherent imperfections of such captions, which often contain errors, typos, and anomalies. These imperfections provide a realistic testbed for evaluating the ability of LLMs to clean and enhance text data. By transforming multimodal memes, which combine text and images, into purely textual data through high-quality captions, we enable the use of NLP techniques alone, reducing the computational complexity associated with multimodal processing \citep{hessel2019unsupervised, lu2019vilbert}. Our work builds upon prior studies by moving beyond GPT models and simpler datasets, evaluating a diverse range of LLMs, including LLaMA 3.1 70B\footnote{\url{https://ai.meta.com/blog/meta-llama-3-1/}}, GPT-4\footnote{\url{https://openai.com/research/gpt-4}}, and SONNET 3.5 v2\footnote{\url{https://www.anthropic.com/news/3-5-models-and-computer-use}}, each offering distinct capabilities and approaches.

The integration of LLMs into data preprocessing pipelines raises important questions about the scalability, reliability, and measurable impact of such approaches. While LLMs demonstrate impressive language understanding capabilities, their application as data preprocessors introduces unique challenges, including the need to maintain semantic consistency while correcting errors and preserving task-relevant information during the cleaning process. These challenges are particularly significant in tasks involving complex, real-world datasets, such as persuasive meme detection, hateful memes, and toxic comments classification, where context plays a crucial role in determining meaning and intent. This stands in contrast to simpler datasets, such as sentiment analysis on short reviews or spam email detection, which often involve limited context and straightforward classification goals. In this study, we focus on persuasive memes as our case study, employing rigorous statistical analysis and empirical evaluation to provide initial insights into both the potential and limitations of LLM-assisted data preprocessing for such tasks.

\begin{table*}[htbp]
\centering
\caption{Caption Coverage Statistics for BLIP and GIT Models Using Sonnet 3.5, LLaMA 3.1 70B, and GPT-4 Turbo}
\resizebox{\textwidth}{!}{%
\begin{tabular}{@{}llccc|ccc@{}}
\toprule
\textbf{Model} & \textbf{Set} & \multicolumn{3}{c|}{\textbf{Non-empty Captions (\#)}} & \multicolumn{3}{c}{\textbf{Valid Captions (\%)}} \\ 
\cmidrule(lr){3-5} \cmidrule(l){6-8} 
               &              & \textbf{Sonnet 3.5} & \textbf{LLaMA 3.1 70B} & \textbf{GPT-4 Turbo} & \textbf{Sonnet 3.5} & \textbf{LLaMA 3.1 70B} & \textbf{GPT-4 Turbo} \\ \midrule
\textbf{BLIP}  & Train        & 6293                & 6993                   & 4844                 & 89.9\%              & 99.9\%                & 69.2\%              \\
               & Dev          & 898                 & 998                    & 726                  & 89.8\%              & 99.8\%                & 72.6\%              \\
               & Test         & 895                 & 1000                   & 718                  & 89.5\%              & 100.0\%               & 71.8\%              \\ \midrule
\textbf{GIT}   & Train        & 5075                & 6755                   & 4872                 & 72.5\%              & 96.5\%                & 69.6\%              \\
               & Dev          & 676                 & 958                    & 698                  & 67.6\%              & 95.8\%                & 69.8\%              \\
               & Test         & 697                 & 976                    & 700                  & 69.7\%              & 97.6\%                & 70.0\%              \\ \bottomrule
\end{tabular}%
}
\label{tab:caption_coverage}
\end{table*}

\section{Method}

The dataset used in this study is from the SemEval 2024 task 4 on hierarchical multi-label classification of persuasive memes. This task involves classifying memes into multiple persuasive techniques organized in a hierarchical structure, comprising 31 distinct labels connected by 36 edges in a directed graph.
The official evaluation metrics for this task are the hierarchical F1 score, accompanied by hierarchical precision and hierarchical recall. These metrics consider the hierarchical relationships among labels, providing a more comprehensive assessment of a model's performance in accurately identifying both the specific and general persuasive techniques present in the memes.

\subsection{Data Preprocessing}
We used the following prompt to clean and correct the captions generated by the BLIP and GIT models:

\begin{quote}
\textit{You are an AI assistant that cleans and corrects image descriptions. Improve the following description by fixing grammatical errors, removing repetitive phrases, and ensuring it is clear and coherent. Provide only the cleaned description without any additional notes or explanations. If the description is too corrupted to fix, respond with 'INVALID DESCRIPTION'.}

\end{quote}

Using this prompt, we processed the captions and discarded any that resulted in an \texttt{INVALID DESCRIPTION}. Consequently, only the meme text was utilized for training when the image caption was invalid. Figure~\ref{fig:caption_cleaning_process} illustrates the workflow of the caption cleaning process for a sample caption generated by the BLIP model.
Table~\ref{tab:caption_coverage} summarizes the performance of the three LLMs in cleaning the captions from the BLIP and GIT models across the training (7,000 images), development (1,000 images), and test (1,500 images) sets.

The results reveal distinctive patterns in how different LLMs approach caption cleaning. \textbf{LLaMA 3.1 70B} demonstrated remarkably high retention rates, maintaining 95--100\% valid captions across all datasets for both BLIP and GIT outputs. This notably high percentage of valid captions suggests a more permissive approach to caption validation. \textbf{Sonnet 3.5} showed varying performance depending on the image captioning model, while it maintained relatively high validity rates for BLIP captions (around 89\%), its performance was lower for GIT outputs (67--72\%). This variation may reflect inherent differences in the quality or characteristics of captions generated by BLIP and GIT, with BLIP outputs potentially offering better quality or being more aligned with \textbf{Sonnet 3.5}'s processing approach. \textbf{GPT-4 Turbo} exhibited the most consistent behavior across both captioning models, with valid caption percentages hovering around 70\%, suggesting it applies a more stringent and uniform validation threshold regardless of the captioning source.

These observations highlight differences in how each LLM handles corrupted captions. While LLaMA 3.1 70B resulted in more valid captions, it is possible that it was less strict in identifying truly invalid descriptions, which could affect the quality of the training data. In contrast, GPT-4 Turbo may have been more conservative, leading to a higher number of captions being marked as invalid and subsequently discarded. This underscores the importance of carefully evaluating the cleaning process to balance the quantity and quality of the data for downstream tasks like multilabel persuasion detection in memes.

\subsection{Training and Model Performance}

\begin{table}[htbp]
\centering
\caption{Performance comparison of BLIP models using uncleaned captions and captions cleaned by different LLMs. Metrics include Precision, Recall, and F1 on both development and test sets, highlighting the effectiveness of LLMs in improving caption quality for downstream tasks. For each LLM, the results correspond to the combination of captions cleaned by that LLM and the original meme text.}
\resizebox{\columnwidth}{!}{%
\begin{tabular}{@{}lcccc@{}}
\toprule
\textbf{LLM}            & \textbf{Set}  & \textbf{Precision} & \textbf{Recall} & \textbf{F1} \\ \midrule
Baseline (No Caption)  & Dev  & 73.95 & 56.72 & 64.20 \\
                        & Test & 67.84 & 47.35 & 55.77 \\ \midrule
Uncleaned Caption       & Dev  & 75.83 & 56.52 & 64.77 \\
                        & Test & 68.47 & 49.03 & 57.14 \\ \midrule
Sonnet 3.5              & Dev  & 74.83 & 58.69 & 65.78 \\
                        & Test & 65.65 & 50.75 & 57.25 \\ \midrule
LLaMA 70B               & Dev  & 73.35 & 57.94 & 64.74 \\
                        & Test & 67.66 & 52.11 & \textbf{58.87} \\ \midrule
GPT-4 Turbo             & Dev  & 74.76 & 58.83 & \textbf{65.85} \\
                        & Test & 69.82 & 49.55 & 57.96 \\ \bottomrule
\end{tabular}%
}
\label{tab:blip_llms}
\end{table}

\begin{table}[htbp]
\centering
\caption{Performance comparison of GIT models using uncleaned captions and captions cleaned by different LLMs. Metrics include Precision, Recall, and F1 on both development and test sets, highlighting the effectiveness of LLMs in improving caption quality for downstream tasks. For each LLM, the results correspond to the combination of captions cleaned by that LLM and the original meme text.}
\resizebox{\columnwidth}{!}{%
\begin{tabular}{@{}lcccc@{}}
\toprule
\textbf{LLM}            & \textbf{Set}  & \textbf{Precision} & \textbf{Recall} & \textbf{F1} \\ \midrule
Baseline (No Caption)  & Dev  & 73.95 & 56.72 & 64.20 \\
                        & Test & 67.84 & 47.35 & 55.77 \\ \midrule
Uncleaned Caption       & Dev  & 73.35 & 57.62 & 64.54 \\
                        & Test & 67.66 & 46.43 & 55.07 \\ \midrule
Sonnet 3.5              & Dev  & 74.04 & 58.40 & 65.30 \\
                        & Test & 65.35 & 52.83 & 58.43 \\ \midrule
LLaMA 70B               & Dev  & 73.33 & 58.98 & 65.37 \\
                        & Test & 67.71 & 51.79 & \textbf{58.69} \\ \midrule
GPT-4 Turbo             & Dev  & 71.60 & 61.29 & \textbf{66.04} \\
                        & Test & 68.71 & 50.79 & 58.41 \\ \bottomrule
\end{tabular}%
}
\label{tab:git_llms}
\end{table}

We employed Google's T5 model for our experiments, leveraging its decoder-based sequence-to-sequence architecture which inherently captures hierarchical relationships in multilabel classification tasks. This architectural choice aligns well with the official hierarchical F1 (h-F1) evaluation metric of the task.

As shown in Tables \ref{tab:blip_llms} and \ref{tab:git_llms}, the baseline model, trained solely on meme text without image captions, achieves F1 scores of 64.20 and 55.77 on the development and test sets respectively. This establishes a foundation for understanding the impact of incorporating image captions. Adding uncleaned BLIP captions leads to slight improvements (dev F1: 64.77, test F1: 57.14), suggesting that even raw image descriptions provide additional useful signal. Similarly, uncleaned GIT captions show modest gains (dev F1: 64.54, test F1: 55.07).

Introducing LLM-cleaned captions yields varying degrees of improvement. For BLIP-generated captions, all three LLMs demonstrate enhanced performance over uncleaned versions. GPT-4 Turbo achieves the highest development set performance (F1: 65.85), while LLaMA 70B leads on the test set (F1: 58.87). Sonnet 3.5 shows consistent improvements but falls slightly behind its counterparts. Similar patterns emerge with GIT-generated captions, where GPT-4 Turbo excels on the development set (F1: 66.04) and LLaMA 70B on the test set (F1: 58.69). 

Looking at the performance across both captioning models, we observe that GPT-4 Turbo achieves the highest development set scores (BLIP: 65.85, GIT: 66.04), while LLaMA 70B shows stronger performance on the test sets (BLIP: 58.87, GIT: 58.69). Overall, the introduction of LLM-cleaned captions provides modest but consistent improvements over both the baseline and uncleaned captions, with gains ranging from 1 to 3.6 F1 points. These results suggest that while LLM-based caption cleaning can enhance performance in this challenging multilabel persuasion detection task, the magnitude of improvement may be task-dependent.

\subsection{Statistical Significance}

We performed per-instance statistical significance testing to assess the impact of using LLM-cleaned captions compared to uncleaned captions for both GIT and BLIP models. The Wilcoxon Signed-Rank Test was employed due to its suitability for paired, non-parametric data, allowing us to compare the per-instance F1 scores between models. To control for the false discovery rate associated with multiple comparisons, we applied the Benjamini-Hochberg correction to adjust the p-values.

The results are presented in Table~\ref{tab:git_significance} for GIT models and Table~\ref{tab:blip_significance} for BLIP models. For the GIT models, none of the LLM-cleaned captions led to statistically significant improvements over the uncleaned captions after adjustment. Specifically, the adjusted p-values for LLaMA 3.1, Sonnet 3.5, and GPT-4 were 0.1291, 0.3735, and 0.1715, respectively all exceeding the significance threshold of 0.05.

In contrast, the BLIP models showed that using GPT-4 Turbo cleaned captions resulted in a statistically significant improvement over uncleaned captions, with an adjusted p-value of 0.0157. This indicates that GPT-4's caption cleaning process effectively enhances the model's performance. The other LLMs, LLaMA 3.1 and Sonnet 3.5, did not show significant improvements for BLIP models, with adjusted p-values of 0.2049 and 0.7047, respectively.

\begin{table}[htbp]
\centering
\caption{Statistical significance results for GIT models, comparing LLM-cleaned captions to uncleaned captions.}
\resizebox{\columnwidth}{!}{%
\begin{tabular}{@{}lccc@{}}
\toprule
\textbf{LLM vs. Uncleaned} & \textbf{P-Value} & \textbf{Adjusted P-Value} & \textbf{Significant} \\ \midrule
LLaMA 3.1                     & 0.0430  & 0.1291  & No  \\
Sonnet 3.5                     & 0.3735  & 0.3735  & No  \\
GPT-4                       & 0.1143  & 0.1715  & No  \\ \bottomrule
\end{tabular}%
}
\label{tab:git_significance}
\end{table}

\begin{table}[htbp]
\centering
\caption{Statistical significance results for BLIP models, comparing LLM-cleaned captions to uncleaned captions.}
\resizebox{\columnwidth}{!}{%
\begin{tabular}{@{}lccc@{}}
\toprule
\textbf{LLM vs. Uncleaned} & \textbf{P-Value} & \textbf{Adjusted P-Value} & \textbf{Significant} \\ \midrule
LLaMA 3.1                      & 0.1366  & 0.2049  & No  \\
Sonnet 3.5                     & 0.7047  & 0.7047  & No  \\
GPT-4                        & 0.0052  & 0.0157  & Yes \\ \bottomrule
\end{tabular}%
}
\label{tab:blip_significance}
\end{table}

\section{Conclusion}

This work explored using LLMs as data preprocessors for refining noisy text data, specifically for multilabel persuasion detection in memes. Our findings show that LLM-cleaned captions improved model performance compared to both baseline models and uncleaned captions, suggesting that LLMs can enhance training data quality beyond just increasing data quantity.
However, different LLMs showed varying approaches raising questions about whether performance gains come from filtering out problematic text or successfully correcting it. Future research should examine which captions different LLMs retain or modify to better understand their preprocessing mechanisms and inform more principled text cleaning strategies.
While the scope of this study was deliberately focused on a specialized task  ``persuasion detection in memes'' due to its complexity and contextual nature, further research is needed to determine how well these findings generalize to other domains and tasks. Testing similar preprocessing approaches across datasets with different noise sources, contextual demands, and content types would provide better insight into the conditions that make LLMs particularly effective as preprocessors. 

\section*{Acknowledgments}
This work is partially supported by the Australian Research Council (ARC) Centre of Information Resilience (Grant No. IC200100022) and by an ARC Future Fellowship Project (Grant No. FT240100022).

\bibliographystyle{ACM-Reference-Format}

\bibliography{aaai25}

\end{document}